\documentclass{article}

\usepackage[preprint]{neurips_2026}

\usepackage[utf8]{inputenc} 
\usepackage[T1]{fontenc}    
\usepackage{hyperref}       
\usepackage{url}            
\usepackage{booktabs}       
\usepackage{amsfonts}       
\usepackage{nicefrac}       
\usepackage{microtype}      
\usepackage{xcolor}         

\usepackage{amsmath}
\usepackage{amssymb}
\usepackage{mathtools}
\usepackage{amsthm}
\usepackage{listings}

\usepackage[most]{tcolorbox}
\usepackage{enumitem}

\usepackage[capitalize,noabbrev]{cleveref}

\theoremstyle{plain}

\theoremstyle{definition}

\theoremstyle{remark}

\usepackage{xcolor}

\usepackage[textsize=tiny]{todonotes}

\lstdefinestyle{skillstyle}{
  basicstyle=\scriptsize\ttfamily,
  frame=single,
  breaklines=true,
  breakatwhitespace=true,
  breakindent=0pt,
  columns=flexible,
  keepspaces=true,
  showstringspaces=false,
  xleftmargin=0pt,
  xrightmargin=0pt,
  aboveskip=6pt,
  belowskip=6pt,
}

\usepackage{pgfplots}
\pgfplotsset{compat=1.18}

\definecolor{darkred}{rgb}{0.6, 0.0, 0.0}
\definecolor{darkgreen}{rgb}{0.0, 0.5, 0.0}

\hypersetup{
    colorlinks,
    linkcolor={red!60!black},
    citecolor={blue!80!black},
}

\title{Chronos: The AI Co-Historian}

\author{%
  Lorenz Hufe$^{*,1,2}$
  \And
  Niclas Griesshaber$^{*,2}$
  \And
  Gavin Greif\ $^{*,2}$
  \And
  Sebastian Oliver Eck$^{2}$
  \AND
  \mbox{Pieter Francois$^{2,5}$ \quad Wojciech Samek$^{1,3,4}$ \quad Christian Schroeder de Witt$^{2}$ \quad  Philip Torr$^{2}$}
  \\ \\
  $^{1}$Fraunhofer HHI \quad $^{2}$University of Oxford  \quad $^{3}$ Technische Universität Berlin \\ $^{4}$ BIFOLD – Berlin Institute for the Foundations of Learning and Data\\ $^{5}$ The Alan Turing Institute \\
  $^{*}$Equal contribution \\
  Correspondence: \\
  \texttt{lorenz.hufe@hhi.fraunhofer.de}}

\begin{document}

\maketitle

\begin{abstract}
AI is increasingly supporting, accelerating, and automating scientific discovery across subjects. Yet, the adoption of AI in historical research remains limited due to the lack of specialised solutions for historians. To change this, we introduce Chronos, an AI Co-Historian designed to support historians. It allows researchers to create and customize research workflows through natural-language interaction and share these as Chronos-Extensions with others. Chronos specifically addresses the need of historians for a tool that is specialised, non-technical, highly customizable, and facilitates extensive task evaluation. As a first extension, we introduce Chronos-Extract, which enables researchers to automate the targeted extraction of information from image scans of historical sources. We benchmark Chronos-Extract on three historical source corpora and find that it achieves high task-accuracy across primary sources spanning three centuries and diverse languages, layouts, and typefaces. Chronos is openly available and ready for historians to use on their own primary and secondary sources.
\end{abstract}

\section{Introduction}
In many areas of AI for Science, progress can be measured with quantitative objectives, such as benchmark performance in machine learning and algorithm discovery, target properties of generated materials, or reaction yield in chemistry \citep{novikov2025alphaevolve,zeni2025generative,boiko2023autonomous}. This makes autonomous scientific discovery attractive, since systems can optimize toward measurable improvements. Historical research is different: its most important contributions can rarely be captured by a single quantitative metric, but must be judged qualitatively by expert historians. We therefore argue that AI for History should prioritise human-AI collaboration over fully autonomous historical research.

Historians visit archives, search for, transcribe, contextualise, and interpret primary sources to build their arguments. These primary sources are the core of historical research. Extracting knowledge from them is not only a technical step but an interpretative process that requires domain-specific expertise. Current AI systems do not yet fully support historians in this process and as a result receive little adoption in their research workflows. 

To increase AI adoption among historians, we introduce \texttt{Chronos}: the first AI Co-Historian. In particular, we make the following three contributions in support of our vision for advancing AI for History:
\newpage
\begin{enumerate}
    \item We introduce \texttt{Chronos}, the first AI Co-Historian that provides a flexible framework for designing, adapting, executing, and sharing AI-assisted research workflows through natural language interaction. \texttt{Chronos} enables historians to partially automate tasks across their research process while remaining human-in-the-loop.
    \item We present \texttt{Chronos-Extract}, a modular \texttt{Chronos-Extension} for constructing structured datasets from digitised primary sources. By shifting from manual pipeline implementation to natural language workflow interaction, \texttt{Chronos-Extract} enables the rapid creation and adaptation of VLM-based extraction pipelines tailored to heterogeneous historical sources.
    \item We evaluate \texttt{Chronos-Extract} on three widely used historical corpora in economic history spanning different languages, layouts, and typefaces, demonstrating strong performance. Our approach achieves $F_1 > 0.99$ on two of the three datasets, matching or exceeding results previously attainable only with corpus-specific pipelines.
\end{enumerate}

The code can be accessed at: \text{\url{https://github.com/ai-historian/chronos}}

\begin{figure*}
    \centering
    \includegraphics[width=.9\linewidth]{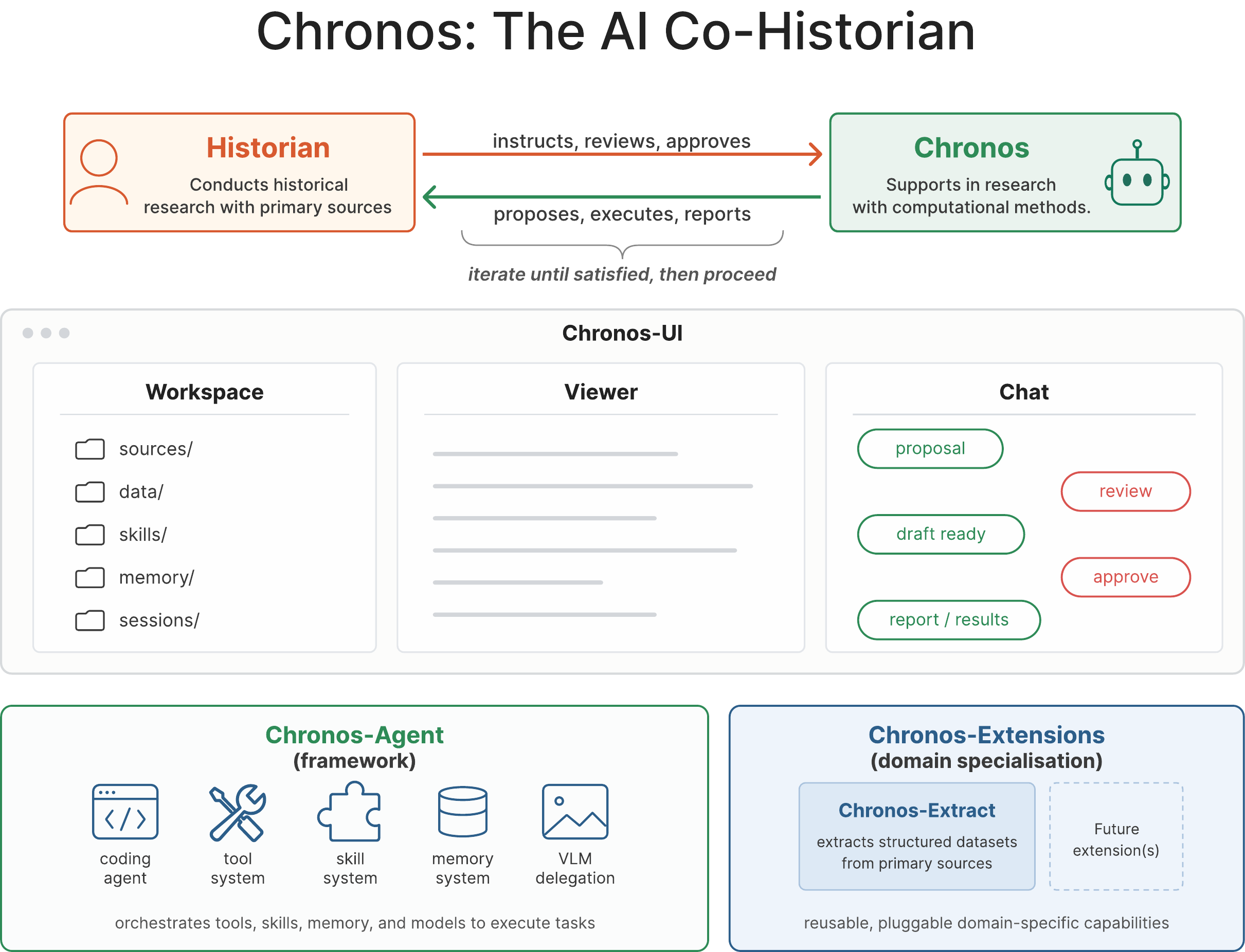}
    \caption{\texttt{Chronos}: an AI Co-Historian that connects historians, digitized primary sources, and reusable computational workflows. Historians interact with the \texttt{Chronos-Agent} through the \texttt{Chronos-UI}, using a dedicated source viewer to inspect evidence, refine instructions, validate outputs, and approve workflows. The agent combines tools, skills, memory, a coding agent, and a VLM subagent to propose, execute, and report computational steps over primary sources. Successful workflows can be packaged as reusable \texttt{Chronos-Extensions}, allowing methods to be shared and adapted across projects. \texttt{Chronos-Extract} is the first such extension and turns scanned primary sources into structured datasets for downstream historical analysis.}
    \label{fig:concept-figure}
\end{figure*}

\section{Related Work}

Extracting information from images is a challenge faced across multiple research domains. In machine learning, a large body of literature addresses the technical challenges of extracting structured data from documents, including optical character recognition (OCR) \citep{fu2024ocrbench,Liu_2024}, document understanding and information extraction \citep{10.1145/3616855.3635752,10.1145/3768156,li2025scoresemanticevaluationframework,li-etal-2025-readoc,Ouyang_2025_CVPR,yang2025cc,zhang2025documentparsingunveiledtechniques}, and reasoning over document content \citep{huang2025ocrreasoningbenchmarkunveilingtrue,Yu_2025_ICCV}.

Building on these advances, researchers in the digital humanities have applied such techniques to historical corpora, creating the data for computational analyses and the exploration of new research questions \citep{Michel_2010,Tyrkko_2022}. Likewise, the social sciences have begun to use these methods to analyse visual data and construct new large-scale economic history datasets \citep{dell2023american,carlson2024efficient,silcock2024newswire,dell2025deep}.

The arrival of multimodal large language models (MLLMs), and in particular vision-language models (VLMs) \citep{liu2023visual}, has further expanded the scope of document extraction. Unlike traditional OCR pipelines, which primarily convert visual text into machine-readable strings, VLMs can jointly perform transcription and semantic parsing, mapping visual document content directly into structured outputs. This enables more flexible and general-purpose extraction workflows, reducing the need for task-specific post-processing pipelines such as named entity recognition or rule-based field extraction.

Many studies using VLMs, however, focus on a single source type or a narrowly defined extraction task. For example, economists have used LLMs to digitise historical tables \citep{backer2025can}, economic historians have constructed datasets from patent registers and firm reports \citep{griesshaber2025multimodal,jayes2025like,xie2025multimodal}, while musicologists have demonstrated that VLMs can extract structured data from structurally complex and diverse sources such as digitised concert programmes \citep{EckPage2026MLLMConcertMetadata}. Recent work has begun to apply (M)LLMs to extract information from diverse museum and archival collections \citep{Schimmenti_2024,Reusens_2025,Toth_2025,vafaie2025end}, and institutions such as the Philadelphia Fed have begun to deploy them on their archival sources \citep{moulton2025harvesting}. 

While these efforts demonstrate the growing applicability of VLMs across diverse historical data sources, prior work typically contributes domain-specific pipelines designed around the layout, notation, and structure of individual corpora. Although effective, such pipelines require substantial manual effort to design and adapt to new sources.

We take a complementary approach: rather than introducing a corpus-specific pipeline, we make extraction pipelines easier for historians to create, customize, and reuse. \texttt{Chronos} enables historians to specify extraction tasks in natural language, which are then translated into VLM-based workflows by the \texttt{Chronos-Agent}. In this sense, \texttt{Chronos-Extract} provides a reusable “pipeline cookbook” that enables the rapid construction and adaptation of extraction pipelines across heterogeneous sources.
 
\section{Chronos: The AI Co-Historian}

\texttt{Chronos} consists of two core components: \texttt{Chronos-Agent}, which accelerates and automates historical research tasks, and \texttt{Chronos-UI}, a user interface that enables historians to supervise and interact with the agent. \texttt{Chronos-Agent} has an extension system that allows users to load, share, and create extensions. This design allows \texttt{Chronos-Agent} to solve domain-specific historical tasks. For example, \texttt{Chronos-Extract} is an extension that specialises \texttt{Chronos-Agent} to support historians in the extraction of structured data from primary sources.

\texttt{Chronos-Agent} is built on pi\footnote{\url{https://github.com/badlogic/pi-mono}}, a general-purpose, model-agnostic coding agent framework. The agent has access to the tools listed in \cref{tab:tools}, which fall into three categories: (a) tools for navigating the host machine's operating system, (b) tools for interacting with \texttt{Chronos-UI}, and (c) tools for querying a VLM subagent.

\texttt{Chronos-UI} builds on the open-source Integrated Development Environment (IDE) Visual Studio Code. Its conceptual design is shown in \cref{fig:concept-figure}, and a screenshot of the implemented interface is provided in Appendix~\cref{app:screenshot}. It is intended to help historians in organising their research environment. \texttt{Chronos-UI} provides functions for (i) initializing the standardised workspace structure (ii) importing and preprocessing primary sources and (iii) starting an agent session. 

\textbf{Skills.} Historians perform many tasks that are highly specific to their chosen topic, region, and period. Skills provide a way to formalise these domain-specific research procedures in natural language \citep{zhang2025skills}, similar to how a historian would explain a task to a collaborator. This improves the performance of \texttt{Chronos-Agent} on such tasks. \texttt{Chronos-Agent} discovers provided skills automatically and applies them if appropriate. \texttt{Chronos-Agent} can also help construct new skills through conversation: the historian describes a task in natural language, the agent proposes a procedure, and this is iteratively refined through human-AI collaboration. Skills are portable, allowing historians to share or adapt procedures across projects. More information about the skill implementation are given in Appendix \cref{app:skills}.

\textbf{Viewer.} \texttt{Chronos-UI} provides a page viewer alongside the conversational interface as shown in \cref{fig:concept-figure}. When the \texttt{Chronos-Agent} analyses a page or the historian asks to inspect one, the scanned image appears in the viewer, allowing the historian to verify the agent's work directly and can be accessed by the agent using the tool described in Table \ref{tab:tools}.

\textbf{Delegated Visual Inspection.}
\texttt{Chronos-Agent} can query information about specific primary source pages without loading their images into its own context by delegating to a non-agentic VLM. This design choice was motivated as we observed significant performance degradation when images were loaded natively into the agent's context. By decoupling visual inspection from the reasoning loop, the agent can choose the model best suited for each task. 

\textbf{Chronos-Extensions.} A \texttt{Chronos-Extension} is a reusable bundle of skills that extends \texttt{Chronos} for a specific domain or task by packaging the procedures, prompts, and  dependencies needed for a specialised workflow.

\begin{table}[t]

    \centering
    \caption{\texttt{Chronos-Agent} tool inventory.}
    \label{tab:tools}
    \small
    \setlength{\tabcolsep}{4pt}
    \begin{tabular}{@{}lll@{}}
      \toprule
      \textbf{Tool} & \textbf{Category} & \textbf{Purpose} \\
      \midrule
      \texttt{read}, \texttt{write}, \texttt{edit}  & OS  & File I/O \\
      \texttt{grep}, \texttt{find}, \texttt{ls}     & OS  & File discovery \\
      \texttt{bash}                                 & OS  & Shell exec \\
      \texttt{list\_pages}                          & OS  & List source pages \\
      \midrule
      \texttt{show\_page}           & UI  & Show page in viewer\\
      \texttt{show\_text}           & UI  & Show text in viewer \\
      \texttt{change\_source}       & UI  & Switch source of viewer \\
      \midrule
      \texttt{analyse\_page}        & VLM & Query VLM \\
      \texttt{follow\_up\_question} & VLM & Continue dialogue \\
      \texttt{analyse\_pages\_batch} & VLM & Batch VLM query \\
      \bottomrule
    \end{tabular}
    
\end{table}

\section{Chronos-Extract: Information Extraction from Historical Sources}

\begin{figure}[t]
    \centering
    \includegraphics[width=\linewidth]{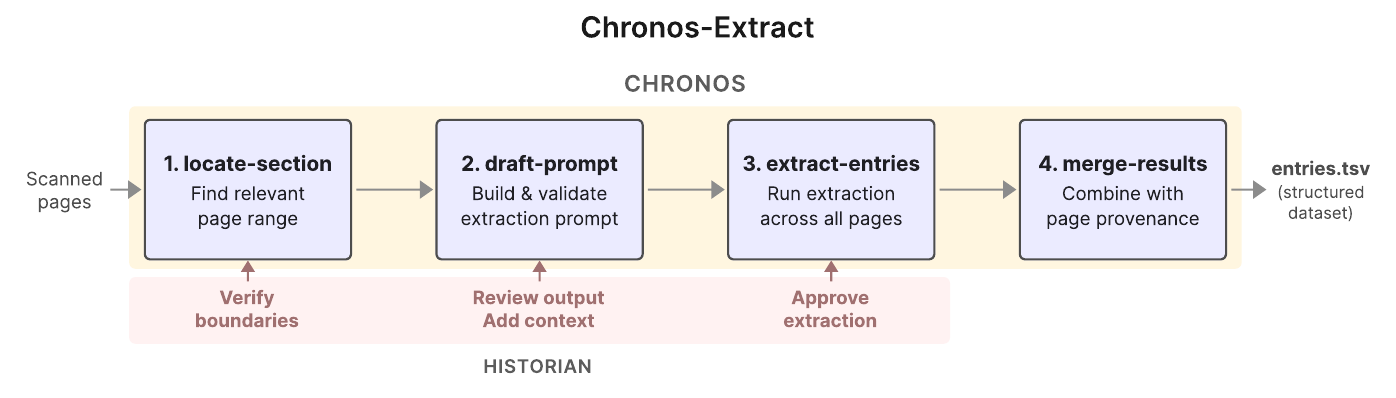}
    \caption{\textbf{\texttt{Chronos-Extract} workflow.} (1) Locate relevant pages, verified by the historian. (2) Draft and interactively refine an extraction prompt. (3) Execute batch extraction across pages after user approval. (4) Merge results into a structured dataset.}
    \label{fig:pipeline}
\end{figure}

\texttt{Chronos-Extract} is a \texttt{Chronos-Extension}, which enables historians to construct custom extraction processes through natural-language interaction with \texttt{Chronos}. \texttt{Chronos-Extract} was developed within \texttt{Chronos}, exclusively through natural language interactions with \texttt{Chronos-Agent}. Below we describe the \texttt{Chronos-Extract} workflow, shown in \cref{fig:pipeline}.

\paragraph{Step 1: Locating the relevant pages.} Historians rarely need an entire primary source, but only the pages relevant to their research question. The first skill (\texttt{locate-section}) addresses this step by employing a binary search over all page images to find the relevant section or, if available, by first consulting the table of contents. To ensure the page range is identified correctly, the \texttt{Chronos-Agent} automatically opens the proposed start and end pages in the page viewer for the historian to verify.

\paragraph{Step 2: Designing the extraction prompt.} Once the relevant pages have been identified, the challenge is to construct an extraction prompt that captures the unique structure and conventions of the primary source. The skill (\texttt{draft-prompt}) searches for structural cues (e.g., abbreviation tables or legends), transcribes them, and incorporates them into a base prompt with source-specific rules and column definitions. \texttt{Chronos-Agent} tests this prompt on a random page using its VLM subagent. Next, the agent presents the extracted entries alongside the original scan to the historian and checks for ambiguities. If issues are identified---for example, that secondary text encodes additional attributes that should be captured in a dedicated column---the agent can revise the prompt until the extraction quality is deemed satisfactory by the historian.

\paragraph{Step 3: Batch extraction.} Before starting the batch extraction \texttt{Chronos-Agent} summarises the page range, displays the final prompt, justifies the model choice, and estimates the cost. Only after the historian's approval does it proceed with the \texttt{ask\_pages\_batch} tool call. 
Once completed, \texttt{Chronos-Agent} examines the results by checking for missing files, column consistency, and common artefacts such as unintended Markdown formatting in the VLM output.

\paragraph{Step 4: Merging with provenance.} For the extracted data to be of value to the historian, the separate files must be concatenated into a dataset suitable for analysis. The final skill (\texttt{merge-results}) performs this step by combining the per-page TSV files and prepending a \texttt{page\_id} column so each row can be traced back to its source page. This step is fully automated and requires no historian input.

\section{Experiments}

To benchmark \texttt{Chronos-Extract}, we evaluate it on three historical source corpora that are both relevant to ongoing historical research and accompanied by gold data ground truth data: German Patent Registers \citep{griesshaber2025multimodal}, French Trade Directories \citep{abadie_das_22}, and American Newspapers \citep{dell2023american}. Together, these corpora span three centuries, several languages and typefaces, and a wide range of document layouts, from structured registers to dense multi-column newspapers. They therefore provide a realistic test bed for evaluating whether \texttt{Chronos-Extract} can generalize across heterogeneous primary sources used in historical research.

For each dataset, we define the target record set as \(Y = \{r_1, \dots, r_N\}\) and the prediction set as \(\hat{Y} = \{\hat{r}_1, \dots, \hat{r}_M\}\). Each record is defined as \(r_i = (t_i, \mathbf{v}_i)\), where \(t_i\) is the full text used for entry-level matching and \(\mathbf{v}_i\) denotes structured variables associated with the record.

\subsection{Datasets}

\textbf{American Newspapers.} Historical newspapers are a valuable source for historians and are a particularly demanding test case for information extraction systems, as a large amount of text is densely packed into a single image scan, typically arranged in multi-column layouts with varying typefaces, print degradation, and article text that spans non-contiguous regions. We evaluate \texttt{Chronos-Extract} on ten fully annotated pages that were created by \citet{dell2023american} to benchmark the accuracy of a large-scale US newspaper dataset they constructed. \cref{fig:app-newspaper} depicts one of the newspaper scans for which annotations exist. Since this dataset does not provide structured variables, we only use the full text entries \(t_i\) for record matching and transcription evaluation.

\textbf{French Trade Directories.} Trade directories are a valuable source for studying urban societies, their occupational structures, and social networks. We evaluate \texttt{Chronos-Extract} on a corpus of 8,765 manually corrected directory entries drawn from 78 pages across 18 French directories with different layouts published between 1798 and 1861. \cref{fig:Bottin1} depicts an image scan of one of the French directory types. \citep{abadie2022dataset, abadie2022benchmark, tual2023benchmark, dumenieu2023entry}. Each directory record contains a \texttt{full\_text} field and six structured tags: \texttt{Person}, \texttt{Activity}, \texttt{Location}, \texttt{House No.}, \texttt{Title}, and \texttt{Facility}.

\textbf{German Patent Registers.} Historians consult patent registers to study innovation and technological change. We evaluate \texttt{Chronos-Extract} on a ground truth dataset consisting of 41 image scans published between 1877 and 1918 by Germany's Imperial Patent Office. Patent registers differ from other historical sources in that they often appear in a structured, double-column list format and include many elements on a page that historians may not wish to extract, such as lists of patent numbers or subcaptions of technological classes. These patent registers were partially printed in Fraktur typefaces, which pose a significant challenge for OCR systems \citep{springmann2018ground}. Each patent record contains a \texttt{full\_entry} field and five structured fields: \texttt{patent\_id}, \texttt{assignee}, \texttt{location}, \texttt{title}, and \texttt{date}.

\begin{figure*}[t]
    \centering
    \includegraphics[width=0.9\linewidth]{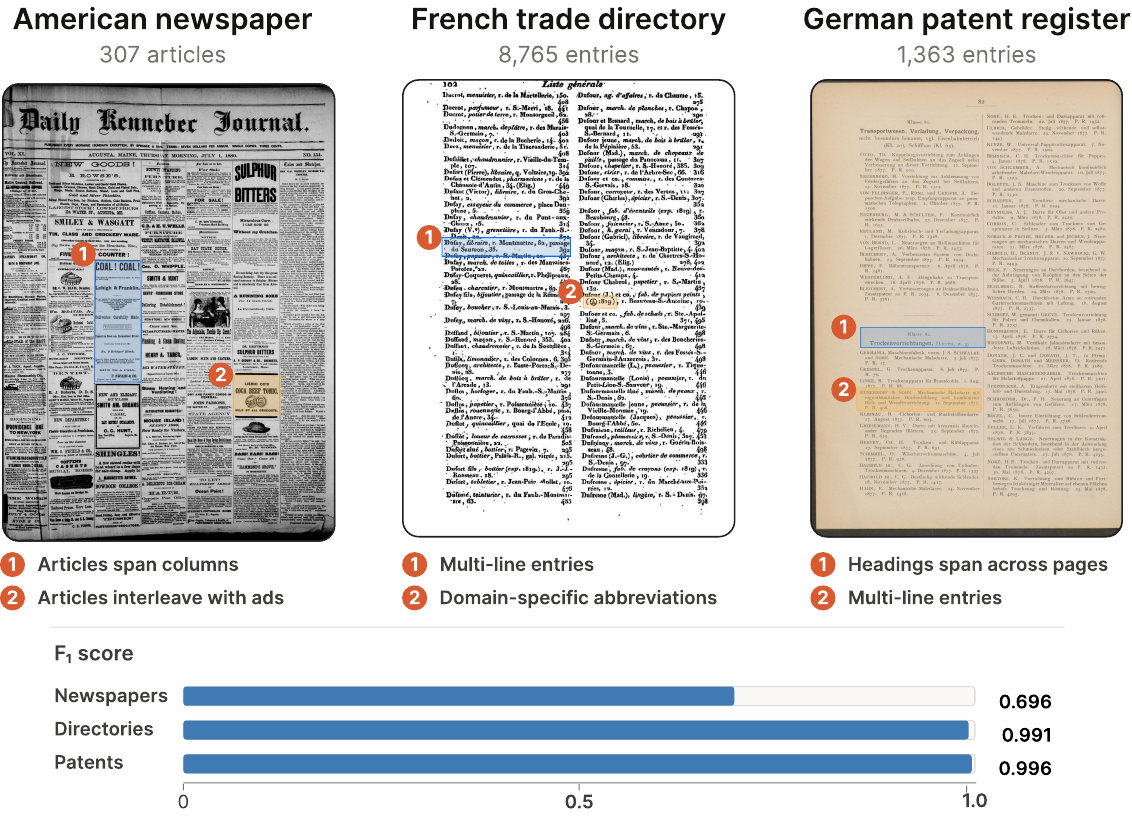}
    \caption{\textbf{Heterogeneity of historical sources.} Example documents from newspapers, trade directories, and patent registers illustrate variation in layout (e.g., multi-column text, interleaved content), encoding conventions (e.g., abbreviations), and typography. These differences require domain-specific interpretation and lead to varying extraction performance (bottom). The source images are shown at high resolution and can be zoomed in for detailed inspection.}
    \label{fig:semantically_complex_information}
\end{figure*}

\subsection{Experimental setup}

We structure the evaluation of \texttt{Chronos-Extract} around the following three questions.

\subsubsection{Does Chronos-Extract identify the records present in the source documents?}
To evaluate whether \texttt{Chronos-Extract} identifies the records present in the source documents, we focus on the full-text fields \(t_i\) and \(\hat{t}_j\). For any two strings \(x\) and \(\hat{x}\), we define the normalised Levenshtein similarity as
\[
\mathcal{S}(x,\hat{x})
=
1 -
\frac{\operatorname{Lev}(x,\hat{x})}{\max(|x|,|\hat{x}|)}.
\]
We compute pairwise similarities \(\mathcal{S}(t_i,\hat{t}_j)\) for all target and predicted records, and use the Hungarian algorithm to obtain the maximum-similarity matching \(A\). We discard all matched pairs below the threshold \(\tau = 0.8\), yielding the set of successful record matches
\[
\mathcal{M}
=
\{(r_i,\hat{r}_j) \mid (i,j)\in A,\ \mathcal{S}(t_i,\hat{t}_j) \geq \tau\}.
\]
Given \(\mathcal{M}\), we compute precision, recall, and \(F_1\). We additionally report the absolute counts of missed records, i.e. unmatched target records, and extra records, i.e. unmatched predictions.

\subsubsection{How accurately does Chronos-Extract transcribe the records? }

To evaluate transcription accuracy, we compute character error rate (CER) on concatenated text fields, following OCR evaluation practice \citep{heidenreich2026gutenocr}. We report two variants. 

First, \(\mathrm{CER}_{\mathcal{M}}\) measures transcription quality on successfully matched records only. Specifically, we concatenate the target texts \(t_i\) for all \((i,j)\in\mathcal{M}\), concatenate the corresponding predictions \(\hat{t}_j\) in the same order, and then compute CER between the two resulting strings.

Second, \(\mathrm{CER}_{\mathcal{C}}\) measures corpus-level transcription quality. Here, we compute CER between the concatenation of all target text fields \(t_1,\dots,t_N\) and the concatenation of all predicted text fields \(\hat{t}_1,\dots,\hat{t}_M\). Unlike \(\mathrm{CER}_{\mathcal{M}}\), \(\mathrm{CER}_{\mathcal{C}}\) also reflects missed records, hallucinated records, and ordering errors.

\subsubsection{How accurately does Chronos-Extract extract structured variables from the records?}
For directories and patents, records include structured variables in addition to their full-text field. We evaluate variable extraction on successfully matched records \((i,j)\in\mathcal{M}\). Let \(v_{ik}\) denote the value of field \(k\) in target record \(r_i\), and let \(\hat{v}_{jk}\) denote the corresponding predicted value in \(\hat{r}_j\). Before comparison, we apply a normalization function \(\nu(\cdot)\) that collapses whitespace and removes trailing punctuation and enclosing quotes.

We report two match rates for each field \(k\). A strict match requires exact equality after normalization,
\[
\nu(v_{ik}) = \nu(\hat{v}_{jk}).
\]
A relaxed match allows minor transcription differences and requires normalized Levenshtein similarity above the same threshold used for record matching,
\[
\mathcal{S}\!\left(\nu(v_{ik}), \nu(\hat{v}_{jk})\right) \geq \tau.
\]
We report strict and relaxed match rates separately for each directory tag and patent field.

\begin{table}[h]
\centering

\begin{minipage}{\linewidth}
\centering
\footnotesize
\caption{Extraction performance by domain and source at matching threshold $\tau = 0.8$. FN (false negative): \(t_i\) entries missed by the prediction. FP (false positive): \(\hat{t}_i\) prediction entries produced by \texttt{Chronos-Extract} absent from ground truth. $\mathrm{CER}_{\mathcal{M}}$: on matched entries only. $\mathrm{CER}_{\mathcal{C}}$: on all of the corpus.}
\label{tab:cross-domain}
\begin{tabular}{@{}lrrrrrr@{}}
\toprule
Source & $N$ & $F_1$ & $\mathrm{CER}_{\mathcal{M}}$ & $\mathrm{CER}_{\mathcal{C}}$ & FN & FP \\
\midrule
\textbf{Patents} & 1363 & $0.996$ & $0.020$ & $0.028$ & 9 & 3 \\
\quad \textit{Antiqua} & 542 & $0.997$ & $0.007$ & $0.013$ & 3 & 0 \\
\quad \textit{Fraktur} & 821 & $0.995$ & $0.029$ & $0.038$ & 6 & 3 \\
\midrule
\textbf{Directories} & 8765 & $0.991$ & $0.020$ & $0.044$ & 23 & 144 \\
\quad Bottin1 & 1480 & $0.994$ & $0.014$ & $0.029$ & 1 & 18 \\
\quad Bottin3 & 872 & $0.989$ & $0.011$ & $0.036$ & 6 & 13 \\
\quad Cambon & 665 & $0.997$ & $0.015$ & $0.034$ & 0 & 4 \\
\quad Deflandre & 738 & $0.970$ & $0.034$ & $0.082$ & 11 & 34 \\
\quad Didot & 2816 & $0.994$ & $0.017$ & $0.046$ & 0 & 35 \\
\quad DidotBottin & 776 & $0.990$ & $0.016$ & $0.027$ & 1 & 14 \\
\quad Duverneuil & 790 & $0.989$ & $0.048$ & $0.062$ & 4 & 14 \\
\quad Favre & 230 & $0.991$ & $0.029$ & $0.042$ & 0 & 4 \\
\quad La~Tynna & 307 & $0.989$ & $0.014$ & $0.048$ & 0 & 7 \\
\quad Notables & 91 & $0.995$ & $0.018$ & $0.028$ & 0 & 1 \\
\midrule
\textbf{Newspapers} & 506 & $0.696$ & $0.018$ & $0.531$ & 201 & 65 \\
\quad Articles & 307 & $0.759$ & $0.018$ & $0.523$ & 102 & 28 \\
\quad Headlines & 199 & $0.595$ & $0.021$ & $0.757$ & 99 & 37 \\
\bottomrule
\end{tabular}
\end{minipage}

\vspace{1em}

\begin{minipage}{\linewidth}
\centering
\footnotesize
\caption{Variable-level match rates on matched entry pairs. \emph{Strict}: exact string match after normalization. \emph{Relaxed}: normalised Levenshtein similarity $\geq 0.8$. $N$: number of aligned variable pairs evaluated.}
\label{tab:variable-accuracy}
\begin{tabular}{@{}llrrr@{}}
\toprule
& Variable & Strict & Relaxed & $N$ \\
\midrule
\textit{Patent fields}
  & ID & $0.999$ & $0.999$ & 1354 \\
  & Person & $0.691$ & $0.968$ & 1354 \\
  & Location & $0.671$ & $0.967$ & 1354 \\
  & Description & $0.374$ & $0.923$ & 1354 \\
  & Date & $0.989$ & $0.999$ & 1354 \\
\midrule
\textit{Directory tags}
  & Person & $0.837$ & $0.909$ & 8742 \\
  & Activity & $0.771$ & $0.900$ & 6533 \\
  & Location & $0.735$ & $0.952$ & 8724 \\
  & House No. & $0.952$ & $0.954$ & 8466 \\
  & Title & $0.053$ & $0.063$ & 917 \\
  & Facility & $0.047$ & $0.049$ & 506 \\
\bottomrule
\end{tabular}
\end{minipage}

\end{table}

\subsection{Results}
We initialize \texttt{Chronos} with Opus 4.6 as the orchestrator and Gemini 3.1-Pro-Preview as the VLM for batch extraction. We use Gemini 3.1-Pro-Preview because it is the strongest Gemini model at the time of writing, and \citet{semnani2025churro} found Gemini models to perform best on historical font understanding. In collaboration with \texttt{Chronos}, we construct one dataset-specific extraction prompt per benchmark on a holdout document to generate our results. The resulting prompts are shown in Appendix \ref{app:prompts}. During the execution of the workflow we auto accepted all tool calls and did not provide additional guidance during the extraction stages.

\textbf{Retrieval quality.}
\texttt{Chronos-Extract} achieves near-perfect record-level retrieval on the German Patent Registers and French Trade Directories, with \(F_1 > 0.99\) on both datasets, but performs substantially worse on American Newspapers, where it reaches \(F_1 = 0.696\) (\cref{tab:cross-domain}). This suggests that language and historical typography are not the main bottlenecks in our evaluation: both German and French sources are retrieved reliably, and the challenging Fraktur subset of the patent registers remains close to the Antiqua subset. Instead, retrieval performance appears to be driven primarily by layout complexity. The structured two-column layouts of patents and directories allow \texttt{Chronos-Extract} to identify records reliably, whereas the dense multi-column newspaper pages contain interleaved articles, headlines, advertisements, and non-contiguous article continuations. These layout features lead to substantially more missed records and lower recall.

\textit{Surprisingly} \texttt{Chronos-Extract} \textit{finds entries missing from the human annotated ground truth.} \cref{tab:cross-domain} details the number of false positives and negatives retrievals. \texttt{Chronos-Extract} achieves excellent entry-level matching on the German Patent Registers and French City Directories, leaving just $32$ of $10{,}128$ target set entries unmatched. Crucially, manual inspection of the $144$ unmatched predicted entries \(\hat{t}\) reveals they are not hallucinations and predominantly valid entries absent from the ground truth, indicating \texttt{Chronos-Extract} identified entries that were missed by human annotators. 

\textbf{Transcription Quality.}
The transcription quality for successfully matched pairs at $\tau \ge 0.8$ is very strong (\cref{tab:cross-domain}). The matched record character error rate ($\mathrm{CER}_{\mathcal{M}}$) averages $0.02$ across all sources. Patents printed in modern Antiqua achieve $\mathrm{CER}_{\mathcal{M}} = 0.007$, while the historical Fraktur fonts raise this slightly to $0.029$ indicating that historical fonts can impact transcription quality. Matched newspaper articles too only show a small error rate of $\mathrm{CER}_{\mathcal{M}} = 0.018$. This indicates the subpar newspaper $F_1$-score is a layout-driven detection failure, rather than a transcription quality concern. 
Corpus-level transcription quality is substantially lower for the American Newspapers, where \(\mathrm{CER}_{\mathcal{C}} = 0.531\), compared to \(0.028\) for patents and \(0.044\) for directories. This large gap is caused by the many record-level misclassifications: missed records, extra predicted records, and ordering errors all contribute directly to \(\mathrm{CER}_{\mathcal{C}}\), even though the successfully matched newspaper records are transcribed accurately.

\textbf{Variable-Level Accuracy.}
\cref{tab:variable-accuracy} reports variable-level accuracy on matched patent and directory entries. For the patent registers, \texttt{Chronos-Extract} achieves consistently strong performance. Patent IDs and dates are extracted almost perfectly, and relaxed match rates exceed \(0.92\) for all fields, including longer description fields where exact string agreement is more sensitive to minor transcription differences. The gap between strict and relaxed scores for person, location, and description fields therefore primarily reflects small textual deviations rather than failures to recover the relevant information.

For the French Trade Directories, performance is high on the main semantic fields: person names, activities, locations, and house numbers all reach relaxed match rates of at least \(0.90\). The main weaknesses are \texttt{Title} and \texttt{Facility}. These fields are substantially harder because they are often encoded by compact, source-specific symbols or abbreviations rather than ordinary alphabetic text, as illustrated in \cref{fig:semantically_complex_information} leading to degraded performance. In the ground truth, \(80.5\%\) of \texttt{Title} spans contain at least one non-ASCII Unicode character, and \(44.2\%\) of \texttt{Facility} spans do so.


\section{Discussion and Limitations}

\texttt{Chronos} suggests that natural-language workflow construction can make computational history more accessible while keeping historians directly involved in task design and validation. Beyond the extraction results reported here, this points toward a broader form of reusable research infrastructure: historical workflows that can be shared, adapted, and improved across projects.

Our evaluation deliberately focused on \texttt{Chronos-Extract} rather than \texttt{Chronos} as a general AI Co-Historian because historical research involves a wide range of interpretative tasks that are difficult to benchmark in its entirety. We therefore evaluated \texttt{Chronos-Extract} as a proxy: structured dataset construction is an important component of computational historical research, but it captures only one part of what historians do. 

Our results show that \texttt{Chronos-Extract} performs well on structured registers and directories, but struggles with complex newspaper layouts. Historical sources remain highly heterogeneous, spanning low-resource or extinct languages, handwriting, damaged scans, marginalia, and domain-specific conventions. Expanding evaluation is difficult because high-quality ground truth requires costly manual annotation, often by multiple annotators \citep{dell2023american}. Moreover, VLMs can hallucinate records, especially in complex layouts, which can corrupt downstream historical datasets. Benchmarking and human validation are therefore indispensable, particularly for low-resource languages where VLM performance remains weaker \citep{liu2026omniocrgeneralistocrethnic}.

Finally, the deployment of AI systems for historical research raises questions of cost, sovereignty, and control. The most capable current VLMs are proprietary and typically require sending digitised primary sources to commercial data centers, which can be problematic for sensitive, restricted, or institutionally governed sources. Their use also creates financial concerns, since higher performance often requires higher spending. However, open-source models such as introduced \citet{semnani2025churro} offer a promising path toward reducing these barriers. If sufficiently capable, they could lower costs, improve transparency, and allow archives, universities, and public institutions to run extraction systems under their own governance.

As historians increasingly outsource dataset construction and other historical tasks to AI systems, they entrust them with shaping the basis on which historical inquiry rests. This could introduce errors, biases, and failure modes methodologically important: AI systems do not merely accelerate historical work, but can influence which records are extracted, omitted, or transformed. For this reason, \texttt{Chronos} is designed around human-in-the-loop workflow construction, where historians specify goals, inspect outputs, and approve workflows. Nevertheless, future AI co-historians will require careful evaluation, transparent validation procedures, and continued reflection on how AI-mediated datasets affect historical knowledge production.

\section{Conclusion}

In this paper, we introduced \texttt{Chronos}, an AI Co-Historian that enables historians to create, customize, and share AI-assisted research workflows through natural-language interaction. As a first extension, \texttt{Chronos-Extract} addresses one of the central bottlenecks in computational history: extracting structured datasets from digitized primary sources. Our experiments show that \texttt{Chronos-Extract} can achieve strong performance across heterogeneous documents spanning three centuries, while remaining flexible enough to be adapted to different source types and research questions. More broadly, \texttt{Chronos} points toward a reusable infrastructure for historical research, where workflows, extensions, and datasets can be shared and improved by the community. By reducing the technical and practical barriers to large-scale source processing, \texttt{Chronos} can expand what historians are able to study, while keeping domain expertise central to the formulation of questions, validation of outputs, and interpretation of evidence.

\newpage

\bibliographystyle{plainnat}
\bibliography{neurips/references}

\newpage

\appendix

\section{Chronos-UI Screenshot}
\label{app:screenshot}

\begin{figure}[h]
    \centering
    \includegraphics[width=\linewidth]{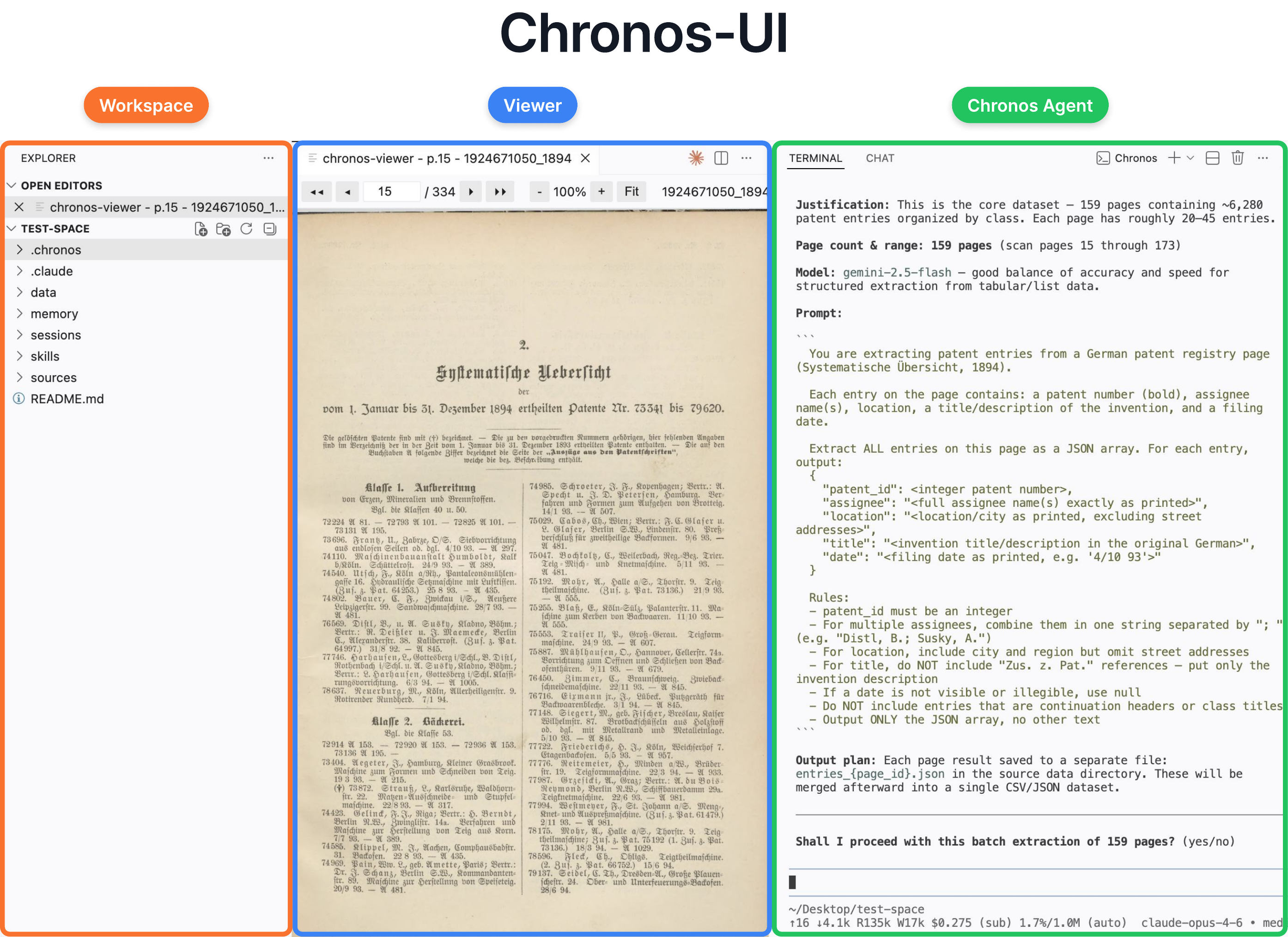}
    \caption{\textbf{\texttt{Chronos-UI}.} Screenshot of the \texttt{Chronos} interface, showing the workspace, source viewer, and conversational agent panel. The interface allows historians to inspect primary sources, interact with the agent, and validate outputs within a single research environment.}
    \label{fig:chronos-ui-screenshot}
\end{figure}

\cref{fig:chronos-ui-screenshot} shows the \texttt{Chronos-UI} during an agent session. The interface combines a standard workspace structure with a conversational agent panel and a source viewer. This allows historians to inspect primary-source scans, review intermediate outputs, and provide feedback without leaving the research environment. The viewer is central to the human-in-the-loop design of \texttt{Chronos}: whenever the agent analyses a source page or proposes a decision, the corresponding scan can be displayed for direct verification by the historian.

\section{Skill Prerequisite Mechanism}
\label{app:skills}

\begin{lstlisting}[basicstyle=\small\ttfamily, frame=single,
  caption={Front matter of a skill that finds the table of contents of a primary source.},
  label={lst:skill}]
---
name: toc-finder
description: Finds the table of contents
requires: none
---
You are an expert analyst of historical
documents (18th-20th century).
 
Step 1 -- Read the per-source memory file...
Step 2 -- Orient: call list_pages...
Step 3 -- Find the Table of Contents...
Step 3.1 -- ...
...
\end{lstlisting}

Each \texttt{SKILL.md} begins with a YAML front matter block containing three fields: \texttt{name} (a human-readable label), \texttt{description} (a one-line summary surfaced in the interface), and \texttt{requires} (a comma-separated list of artefact files that must exist before the skill will run). Listing~\ref{lst:skill} shows a minimal example.

The \texttt{requires} field is the mechanism that turns independent skills into ordered pipelines. When the agent is asked to execute a skill, it checks for each listed file before proceeding and halts with a clear message if any is missing. For instance, the \texttt{batch-extract} skill in the pipeline shown in \cref{fig:pipeline} declares \texttt{requires:~batch\_prompt.md}, ensuring that the \texttt{prompt-construction} skill has completed successfully and produced its artefact before batch extraction begins. This lightweight dependency mechanism enables multi-step workflows without requiring a separate orchestration layer: the historian or agent simply invokes skills in sequence, and the prerequisite check enforces the correct ordering.

\section{Dataset Links}
\label{app:dataset_links}

Table~\ref{tab:dataset_links} provides links to the source collections and annotated datasets used in our evaluation.

\begin{table}[h]
\centering
\caption{\textbf{Links to benchmark datasets.} The table lists the public source collections and ground truth datasets used to evaluate \texttt{Chronos-Extract}.}
\label{tab:dataset_links}
\begin{tabular}{lll}
\toprule
\textbf{Dataset} & \textbf{Link type} & \textbf{Link} \\
\midrule
German Patent Registers 
& Digitized source collection 
& \href{https://digi.bib.uni-mannheim.de/werksansicht/82403/1/einzelseiten/nogridview?cHash=ccbe10e4d5fe2653fc21b6678fdc3fcf}{Mannheim University Library} \\

French Trade Directories 
& Ground truth dataset 
& \href{https://zenodo.org/records/6394464}{Zenodo record} \\

American Newspapers 
& Large-scale source dataset 
& \href{https://huggingface.co/datasets/dell-research-harvard/AmericanStories}{American Stories} \\

American Newspapers 
& Fully annotated evaluation pages 
& \href{https://huggingface.co/datasets/dell-research-harvard/AmericanStoriesTraining}{American Stories Training} \\
\bottomrule
\end{tabular}
\end{table}

\section{Example Skill: \texttt{locate-section}}
\label{app:skill-example}

Listing~\ref{lst:skill-full} shows an adapted version of the \texttt{locate-section} skill for a German historical source, used as Step~1 of the extraction pipeline in \cref{fig:pipeline}. The skill is a single Markdown file that the agent reads and executes step by step; there is no accompanying code. It illustrates the level of detail at which domain procedures are captured: operational constraints (one page at a time, binary search discipline), source-specific knowledge (Berichtigungen and Änderungen as named section types, the printed-vs-tool page ID distinction), an embedded prompt for the vision subagent, and a human-verification loop that gates the final save. Other skills in the pipeline follow the same structural template.

\begin{tcolorbox}[
  breakable,
  colback=white,
  colframe=black,
  boxrule=0.5pt,
  arc=0pt,
  left=4pt, right=4pt, top=4pt, bottom=4pt,
  title={Listing 2: Complete text of the \texttt{locate-section} skill.},
  fonttitle=\small,
  coltitle=black,
  colbacktitle=white,
]
\label{lst:skill-full}
\begin{lstlisting}[
  basicstyle=\scriptsize\ttfamily,
  frame=none,
  breaklines=true,
  columns=flexible,
  keepspaces=true,
  showstringspaces=false,
  aboveskip=0pt, belowskip=0pt,
]
---
name: locate-section
description: Locate the page range of the alphabetical name list in a historical German city directory, plus any Berichtigungen or Aenderungen sections.
---
You are an expert analyst of historical German documents (18th-20th century).

## Setup
The source directory contains PNG scans accessed via tools. To preserve your context, you do not view pages yourself -- delegate to `analyze_page`, a specialist vision agent that returns summaries or answers questions about a single page.
The only valid end state is a verified result file written to the source's `data/` directory. If uncertain about any finding, resolve it with additional tool calls before writing.

## Rules
**One page at a time.** Never issue concurrent `analyze_page` calls. Fire one call, wait, reason, then pick the next page.
**Inspect before reporting.** Every page number you report must be one you personally viewed via `analyze_page`. Never trust a table of contents without verification.
**Page ID != printed page number.** Tool page IDs are sequential file indices. Printed page numbers inside the scan are different (e.g. printed "p. 347" may be tool ID 412). Always use tool IDs in tool calls and in the output.
**Gaps are normal.** Scans may have missing pages or non-consecutive IDs. ID N+1 is not necessarily physically adjacent to ID N. Widen your interval if a boundary search returns unexpected content.
**Verify both sides of every boundary.**
- Start: candidate page contains name-list entries; the page before does not.
- End: candidate page is still name-list; the page after is not.
**Strict binary search.** Given `[lo, hi]` where `lo` is outside and `hi` is inside the section, compute `mid = (lo + hi) // 2`, view it, and narrow. Repeat until `hi - lo == 1`. Do not guess multiple adjacent pages in parallel.
**Memory discipline.** Read `DOCUMENT_MEMORY.md` at the start; it may contain findings from prior runs. Update it every ~10 steps so partial results survive interruption. Prefer `edit` over `write`. Keep it concise.

## Task
Find the page range of the main alphabetical name list (residents, citizens, or subscribers) and identify any Berichtigungen (corrections) or Aenderungen (amendments).
**First occurrence only.** Some directories repeat the name list in multiple orderings (e.g. by surname, then by street). Report only the first. Use the alphabetical ordering itself to detect when a second pass begins.
**Berichtigungen & Aenderungen** may precede, follow, or be embedded within the main list. Search actively -- null is only acceptable after a targeted search, not by assumption.

## Procedure
**Step 1 -- Read memory.** Read `DOCUMENT_MEMORY.md` for prior findings.
**Step 2 -- Orient.** Call `list_pages`. Note first and last page IDs.
**Step 3 -- Table of contents and offset.** Scan the first ~15 pages with `analyze_page` to locate the Inhaltsverzeichnis. If found, read the printed page number for the start of the name list, then find the corresponding tool page ID. Compute `offset = tool_page_id - printed_page_number` and use it to estimate the end boundary from the TOC.
Use this prompt for every `analyze_page` call during search:
> Classify and summarise this page for the purpose of finding the name list in a historical German city directory.
> Step 1 -- Classify: assign one category:
>   A) Listing of names, addresses, or personal information
>   B) Table of contents
>   C) Other (advertisements, narrative text, notices, preface, etc.)
> Step 2 -- Report:
>   - A: state in the first sentence that the page contains a name/address listing. Specify the ordering. Give the first and last entry verbatim. List fields.
>   - B: state in the first sentence that the page is a table of contents. Report it completely, preserving structure and printed page numbers.
>   - C: state in the first sentence that the page does NOT contain a name listing. Describe the content.
**Step 4 -- Coarse scan (fallback).** If the TOC is missing or unusable, call `analyze_page` every ~50 pages to bracket the section manually.
**Step 5 -- Binary search.** For each boundary, bracket `[lo, hi]` and perform strict binary search at the midpoint. No parallel guesses. Stop when one side is confirmed outside and the other inside.
**Step 6 -- Confirm.** Re-run `analyze_page` on both pages of each boundary pair. If anything looks off, return to Step 5.
**Step 7 -- Berichtigungen / Aenderungen.** Targeted search: 3 pages immediately before the start boundary, 3 pages after the end boundary, the first 5 pages, and the last 5 pages.
**Step 8 -- Update memory.** Write a concise summary to `DOCUMENT_MEMORY.md` (page ranges, document structure, anything useful for future runs). Preserve existing content.
**Step 9 -- Save.** Write `name_ranges.json` to the source's `data/` directory using the schema below, with `human_verified_start` and `human_verified_end` set to `false`. Proceed to Step 10.
**Step 10 -- Human verification.** For each boundary in turn:
  1. Call `show_page` with the boundary page ID.
  2. Ask the user: "This is the [first/last] page of the name list (page X). Does this look correct?"
  3. Wait for the user's reply -- do not proceed without it.
  4. If confirmed: mark the corresponding `human_verified_*` field `true`. If rejected: ask follow-up questions, adjust the boundary, re-save, and repeat until confirmed.
  5. After both boundaries are confirmed, save once more with the verified flags. This final save is the only acceptable end state.

## Output schema
Write `name_ranges.json` to the source's `data/` directory:
{
  "source": "<source folder name>",
  "name_list": {
    "start_page":  <int>,
    "end_page":    <int>,
    "description": <string>,
    "human_verified_start": <bool>,
    "human_verified_end":   <bool>
  },
  "amendments": {
    "berichtigungen": {
      "start_page":  <int | null>,
      "end_page":    <int | null>,
      "description": <string | null>
    },
    "aenderungen": {
      "start_page":  <int | null>,
      "end_page":    <int | null>,
      "description": <string | null>
    }
  },
  "notes": <string>
}
All page numbers are tool page IDs. Use `null` for sections actively searched and not found.
\end{lstlisting}
\end{tcolorbox}

\newpage
\section{Prompts}
\label{app:prompts}

\begin{figure}[ht]
    \centering
    \includegraphics[width=1\linewidth]{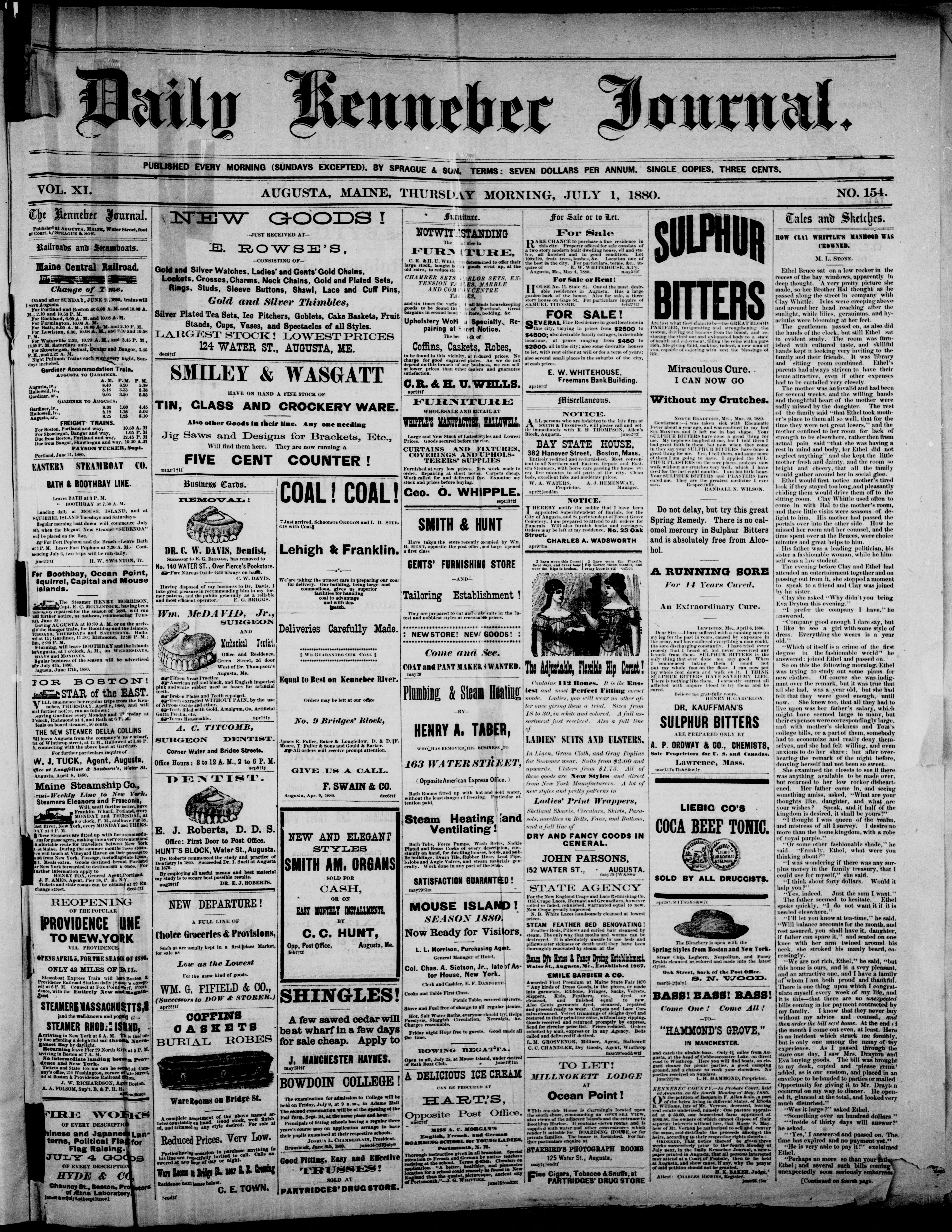} 
    \caption{Sample page from the American Historical Newspapers dataset.}
    \label{fig:app-newspaper}
\end{figure}

This section provides prompts and skills used by \texttt{Chronos-Extract}. They illustrate how extraction workflows can be specified in natural language and adapted to different historical source types. Each prompt encodes domain-specific rules, output formats, and constraints that guide the VLM during extraction.
\begin{tcolorbox}[
  breakable,
  colback=white,
  colframe=black,
  boxrule=0.5pt,
  arc=0pt,
  left=4pt, right=4pt, top=4pt, bottom=4pt,
  title={Listing 3: Extraction prompt (\texttt{benchmarking/prompts/american-newspapers-v4.md}) for historical newspapers (\cref{fig:app-newspaper}).},
  fonttitle=\small,
  coltitle=black,
  colbacktitle=white,
]
\label{lst:newspaper-prompt}
\begin{lstlisting}[
  basicstyle=\scriptsize\ttfamily,
  frame=none,
  breaklines=true,
  columns=flexible,
  keepspaces=true,
  showstringspaces=false,
  aboveskip=0pt, belowskip=0pt,
]
# American Newspapers — Extraction Prompt

Extract every **headline** and **article body** from this historical newspaper page.

## What to extract

- **Headlines**: The title text above an article. Multi-line headlines are common. Do not include bylines or author names.
- **Articles**: The body text of news reports, editorials, correspondence columns, poems, fiction, obituaries, and feature pieces. Start with the first sentence of the body — do not include the headline or byline.

Use model `gemini-3.1-pro-preview` with batch processing.

## What to exclude

Advertisements, classifieds, legal notices, mastheads, banner titles, subscription terms, section headers (e.g., "LOCAL NEWS"), page numbers, and decorative elements.

## Formatting rules

1. **Preserve column line breaks** exactly as printed. Do not reflow text or join hyphenated words across lines.
2. **Preserve original spelling and punctuation.** Do not modernize or correct.
3. **Continuation markers** (e.g., `[Continued on fourth page.`) should be copied as printed.

## Grouping rules

- **Correspondence columns** (a single correspondent reporting multiple local items in one column segment): bundle as **one article**.
- **"Locals" / "personal mentions" pages** where each item has a typographic separator: each item is its **own article**.
- **When in doubt, prefer bundling** over splitting.

## Output format

Return a JSON object with an `entries` array. Each entry has a `type` ("headline" or "article") and `text`.

```json
{
  "entries": [
    {"type": "headline", "text": "Hotel Life in New York."},
    {"type": "article", "text": "Mr. Fitzwalter was the pastor of an\nimpoverished but genteel colored con-\ngregation in Port Royal."},
    {"type": "headline", "text": "The Arrest of Parnell-Details of the\nAffair."},
    {"type": "article", "text": "Charles Stewart Parnell was arrested\nat his hotel in Morrison's..."}
  ]
}
```

## Final output

Consolidate results into a single `.xlsx` file with columns: `page`, `type`, `text`. One row per entry.
\end{lstlisting}
\end{tcolorbox}

\newpage

\begin{figure}
    \centering
    \includegraphics[width=0.9\linewidth]{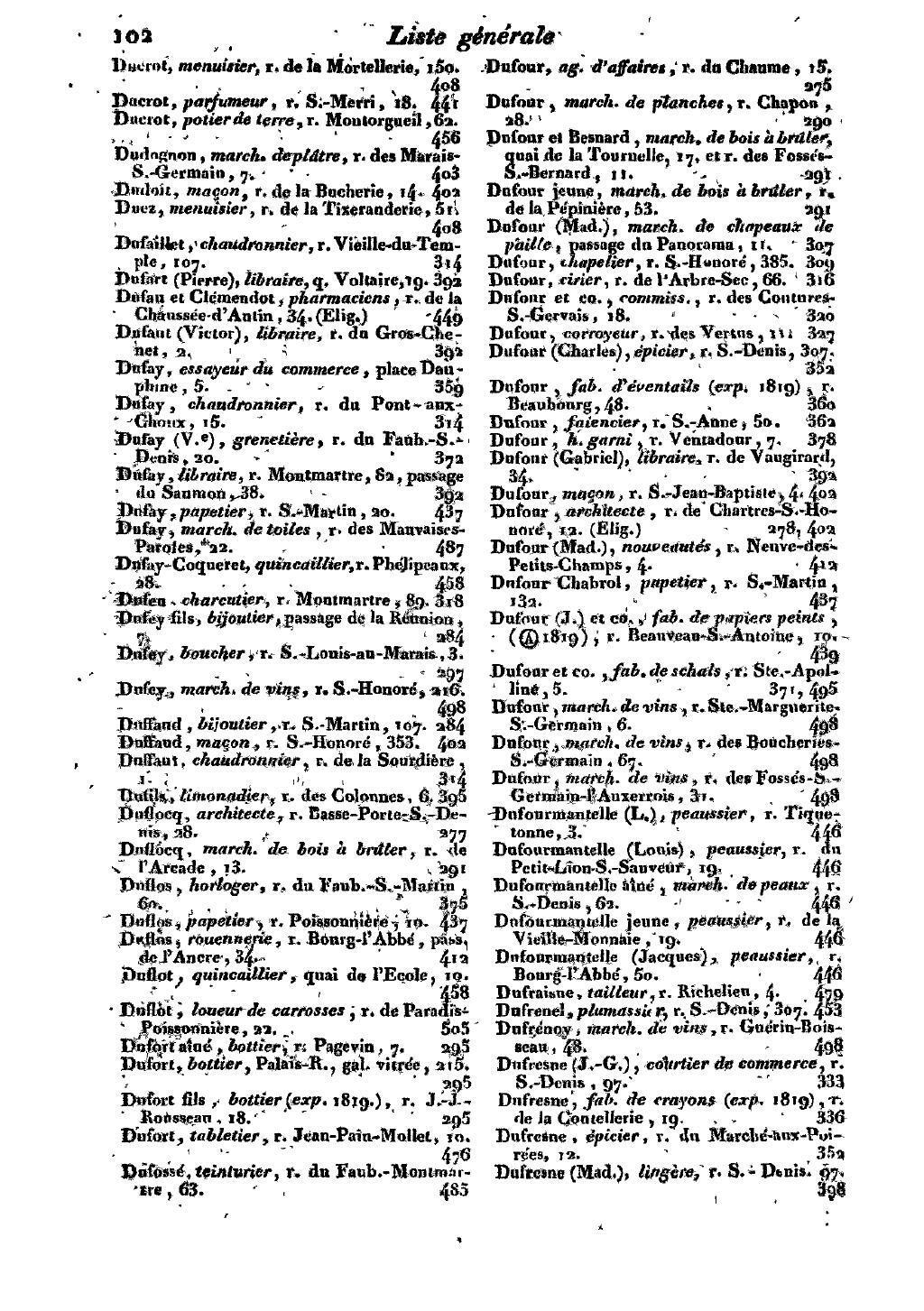}
    \caption{Sample page from French Trade Directories.}
    \label{fig:Bottin1}
\end{figure}

\begin{tcolorbox}[
  breakable,
  colback=white,
  colframe=black,
  boxrule=0.5pt,
  arc=0pt,
  left=4pt, right=4pt, top=4pt, bottom=4pt,
  title={Listing 4: Extraction prompt (\texttt{benchmarking/prompts/french-directories.md}) for French trade directories (\cref{fig:Bottin1}).},
  fonttitle=\small,
  coltitle=black,
  colbacktitle=white,
]
\label{lst:directory-prompt}
\begin{lstlisting}[
  basicstyle=\scriptsize\ttfamily,
  frame=none,
  breaklines=true,
  columns=flexible,
  keepspaces=true,
  showstringspaces=false,
  aboveskip=0pt, belowskip=0pt,
]
# French Trade Directories — Extraction Prompt

First, review the full set of directory page images to familiarize yourself with the layout, typography, and entry structure before extracting.

Use model `gemini-3.1-pro-preview` with batch processing.

Extract every individual business or person entry from this directory page.

Each entry typically contains some combination of: a person or business name, their profession or activity, a street address, and a house number. Some entries also include honorary titles or facility type information.

## Output Format

Return a JSON object with an `entries` array. For each entry, provide:

- `full_text`: The complete entry as continuous plain text. Preserve original spelling, abbreviations, and punctuation. Do not preserve column line breaks — join lines into flowing text.
- `PER`: Array of person or business name strings found in the entry.
- `ACT`: Array of profession or activity strings (e.g., "pharmaciens", "bottier", "essayeur du commerce").
- `LOC`: Array of location or street name strings (e.g., "r. de la Chaussée-d'Antin", "place Dauphine").
- `CARDINAL`: Array of house/street number strings (e.g., "34", "12").
- `TITRE`: Array of title or honor strings (e.g., "(Elig.)", "(Jur.)").
- `FT`: Array of facility type strings (e.g., "fab. à Caen", "dépôt", "boutique").

If an entity type is not present in an entry, use an empty array `[]`.

## Guidelines

- Extract every entry on the page, even short or partially legible ones.
- Preserve original 19th-century French spelling and abbreviations exactly as written (e.g., "r." for "rue", "fab." for "fabrique", "pl." for "place", "V.e" for "Veuve").
- Entries may span multiple lines — combine them into one entry.
- Do not modernize spelling, expand abbreviations, or correct apparent OCR-like artifacts in the original print.
- Some entries may contain numbers at the very end that are reference numbers, not street numbers — include them in `full_text` but only tag actual street/house numbers as `CARDINAL`.

## Example Output

```json
{
  "entries": [
    {
      "full_text": "Dufan et Clémendot, pharmaciens, r. de la Chaussée-d'Antin, 34. (Elig.) 449",
      "PER": ["Dufan et Clémendot"],
      "ACT": ["pharmaciens"],
      "LOC": ["r. de la Chaussée-d'Antin"],
      "CARDINAL": ["34"],
      "TITRE": ["(Elig.)"],
      "FT": []
    }
  ]
}
```

## Final Output

After extracting from all pages, consolidate the results into a single `.xlsx` file for the entire source.
\end{lstlisting}
\end{tcolorbox}

\newpage

\begin{figure}
    \centering
    \includegraphics[width=0.9\linewidth]{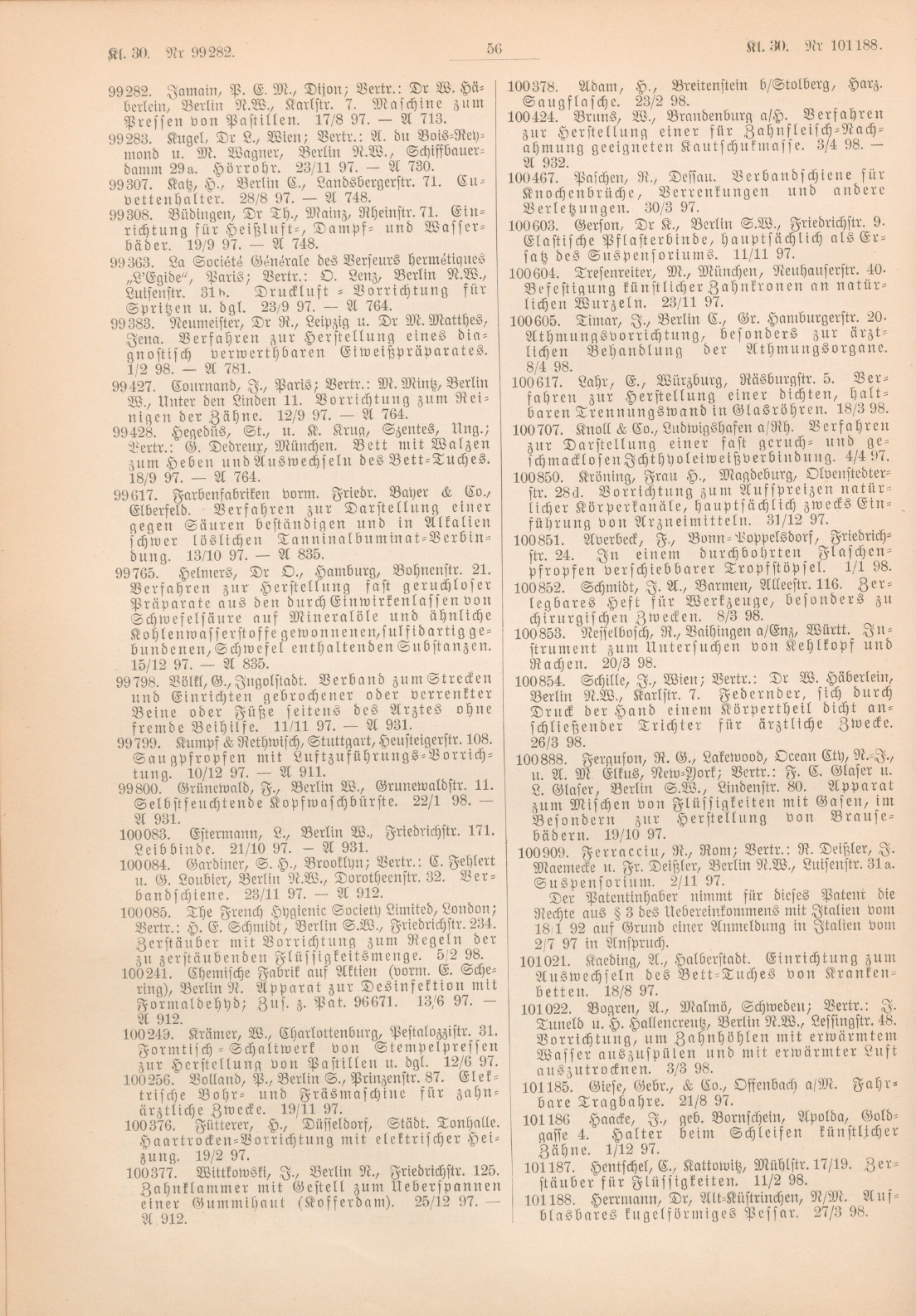}
    \caption{Sample pages from the German Imperial Patent Gazette dataset.}
    \label{fig:patents}
\end{figure}

\begin{tcolorbox}[
  breakable,
  colback=white,
  colframe=black,
  boxrule=0.5pt,
  arc=0pt,
  left=4pt, right=4pt, top=4pt, bottom=4pt,
  title={Listing 5: Extraction prompt (\texttt{benchmarking/prompts/german-patents.md}) for German patent registers (\cref{fig:patents})},
  fonttitle=\small,
  coltitle=black,
  colbacktitle=white,
]
\label{lst:patent-prompt}
\begin{lstlisting}[
  basicstyle=\scriptsize\ttfamily,
  frame=none,
  breaklines=true,
  columns=flexible,
  keepspaces=true,
  showstringspaces=false,
  aboveskip=0pt, belowskip=0pt,
]
# German Patents — Extraction Prompt

First, review the full set of patent page images to familiarize yourself with the layout, typography, and entry structure before extracting.

Use model `gemini-3.1-pro-preview` with batch processing.

Extract every patent entry from this page of the German Patent Office records (Kaiserliches Patentamt / Reichspatentamt).

Each entry typically contains: a patent identification number, the name of the inventor or assignee, optionally a location (city or address), a description of the invention, and a date.

## Output Format

Return a JSON object with a `patents` array. For each entry, provide:

- `full_entry`: The complete entry as continuous plain text. Preserve original spelling and punctuation. Do not preserve column line breaks — join lines into flowing text.
- `patent_id`: The patent number as a string (e.g., "919", "1312").
- `assignee`: The inventor or applicant name exactly as written (e.g., "OTTO, TH.", "PHILIPPE, E., und J. M. GENET").
- `location`: The city or address if present, otherwise `null`. (e.g., "Berlin S.W., Kochstraße 10", "Paris", or `null`).
- `title`: The patent title or description of the invention.
- `date`: The date as printed (e.g., "12. September 1877").

## Example Output

```json
{
  "patents": [
    {
      "full_entry": "OTTO, TH. Kuppelungsvorrichtung zum Anhängen der Wagen auf Seilbahnen an das Zugseil nebst Verbesserung an dieser Vorrichtung. 12. September 1877. P. R. 919.",
      "patent_id": "919",
      "assignee": "OTTO, TH.",
      "location": null,
      "title": "Kuppelungsvorrichtung zum Anhängen der Wagen auf Seilbahnen an das Zugseil nebst Verbesserung an dieser Vorrichtung",
      "date": "12. September 1877"
    }
  ]
}
```

## Guidelines

- Only extract patent entries that are **complete** on the page. Skip entries that are truncated at the top-left or bottom-right of the image (partial entries that continue from a previous or onto a next page).
- Preserve German characters exactly as written (ä, ö, ü, ß, ſ).
- Preserve original spelling and technical terminology — do not translate or modernize.
- Some entries have multiple inventors — include all names in the `assignee` field as a single string.
- The patent ID may appear in different positions within the entry (beginning, end, or prefixed with "P. R.", "D. R. P.", "Nr.", etc.) — extract just the number.

## Final Output

After extracting from all pages, consolidate the results into a single `.xlsx` file for the entire source.
\end{lstlisting}
\end{tcolorbox}

\section{Primary Source Examples}
\label{app:primary_sources}

This appendix provides visual examples of the diverse primary sources used in our benchmarking evaluation. Because layout, typography, and image degradation strongly influence extraction performance, these examples illustrate the complexity and heterogeneity of the historical datasets.

\clearpage

\clearpage

\clearpage
\newpage

\end{document}